\documentclass{article}

\PassOptionsToPackage{numbers, sort&compress}{natbib}

\usepackage[final]{unireps_2023}

\usepackage[utf8]{inputenc} 
\usepackage[T1]{fontenc}    
\usepackage{hyperref}       
\usepackage{url}            
\usepackage{booktabs}       
\usepackage{amsfonts}       
\usepackage{nicefrac}       
\usepackage{microtype}      
\usepackage{xcolor}         
\usepackage{mathtools}
\usepackage{amssymb}
\usepackage{graphicx}
\usepackage{algpseudocode}
\usepackage{caption}
\usepackage{subcaption}
\usepackage{multirow}
\usepackage{multicol}
\usepackage{tablefootnote}
\usepackage{wrapfig}
\usepackage{textcomp}
\usepackage{siunitx}
\usepackage{multirow}
\usepackage{wrapfig}
\usepackage{booktabs}
\usepackage{graphics}
\usepackage{multibib}
\usepackage{gensymb}
\usepackage[shortlabels]{enumitem}

\newtheorem{define}{Definition}
\newtheorem{assum}{Assumption}

\title{Randomly Weighted Neuromodulation in Neural Networks Facilitates Learning of Manifolds Common Across Tasks}

\author{%
  Jinyung Hong$^{1}$ \quad\quad Theodore P.~Pavlic$^{1, 2}$ \\
  \textsuperscript{1}School of Computing and Augmented Intelligence \\
  \textsuperscript{2}School of Life Sciences \\
  Arizona State University \\
  Tempe, AZ 85281 \\
  \texttt{\{ jhong53, tpavlic \}@asu.edu} \\
}

\begin{document}

\maketitle

\begin{abstract}
Geometric Sensitive Hashing functions, a family of Local Sensitive Hashing functions, are neural network models that learn class-specific manifold geometry in supervised learning. However, given a set of supervised learning tasks, understanding the manifold geometries that can represent each task and the kinds of relationships between the tasks based on them has received little attention. We explore a formalization of this question by considering a generative process where each task is associated with a high-dimensional manifold, which can be done in brain-like models with neuromodulatory systems. Following this formulation, we define \emph{Task-specific Geometric Sensitive Hashing~(T-GSH)} and show that a randomly weighted neural network with a neuromodulation system can realize this function.\footnote{All source codes and figures are available at \url{https://github.com/PavlicLab/NeurIPS2023-UniReps-Hong-TGSH-Randomly_Weighted_Neuromodulation_in_Neural_Networks.git}.}
\end{abstract}

\section{Introduction}
\label{sec:introduction}
Deep Neural Networks~(DNNs) have shown outstanding performance in various fields due to their ability to learn to transform complex objects into useful representations that can be easily separated in the embedding space. However, due to their ``black box'' behavior, there is still little known about exactly which information-rich features are captured by a DNN representation and what information-poor variations are eliminated. To address the question, \citet{dikkala2021manifold} considered the manifold geometry of a generative process where each class is associated with a manifold and parameter that shifts class examples along that manifold. Using this perspective, they could show that neural representations in supervised learning could be viewed as a kind of Locality Sensitive Hash functions that they coined Geometric Sensitive Hashing~(GSH)~\cite{dikkala2021manifold}.

In this work, we extend this geometric interpretation from one classification task to a series of potentially related classification tasks. For example, in a video clip of digits rotating counterclockwise, each possible digit within a single frame is a class associated with its own manifold, as in GSH, and we focus on the relationship between the manfolds induced by the classification problem in one frame (i.e., one rotation angle) and the other frames. Such a manifold-learning perspective allows us to compare representations of tasks that are similar in principle so that we can identify representational kernels that are shared among each individual classification task. In essence, for this example digit-rotation task, we consider the possibility that rotated digit images may in fact map to different points on the same induced digit manifold that is then combined with a rotation manifold to represent the image and its orientation.



The geometry of low- and high-level perceptual spaces has been studied in cognitive science, and changes in those behavioral and cognitive states have been associated with neuromodulatory systems in the brain. Neuromodulation has been studied extensively in human and insect brains, where neuromodulatory signals act as a kind of \emph{switchboard} that can remap a neural representation in different ways so as to include adaptive control of behavior based on internal state and environmental context~\cite{modi2020drosophila, mei2022informing}. In response to the importance of neuromodulatory systems in biological brains, neuromodulation-inspired neural networks have been proposed in various domains, such as reinforcement learning~\cite{botvinick2020deep}, modular networks~\cite{daram2020exploring}, and adaptive behaviors~\cite{vecoven2020introducing}. For example, promising applications of neuromodulated neural networks to the problem of to Continual Learning~(CL)~\cite{daram2020exploring, hong2022learning, miconi2020backpropamine} can prevent catastrophic forgetting by modulating synapsis that remain persistently stable.


In this work, we connect neuromodulation-inspired neural networks for continual learning and geometric manifold learning in supervised learning. We introduce the definition of \emph{task-specific manifolds} and demonstrate that DNNs with neuromodulators can learn these manifolds.  

%
\begin{figure*}[t!]\centering%
\begin{subfigure}[t]{0.13\textwidth}\centering
\includegraphics[width=\textwidth]{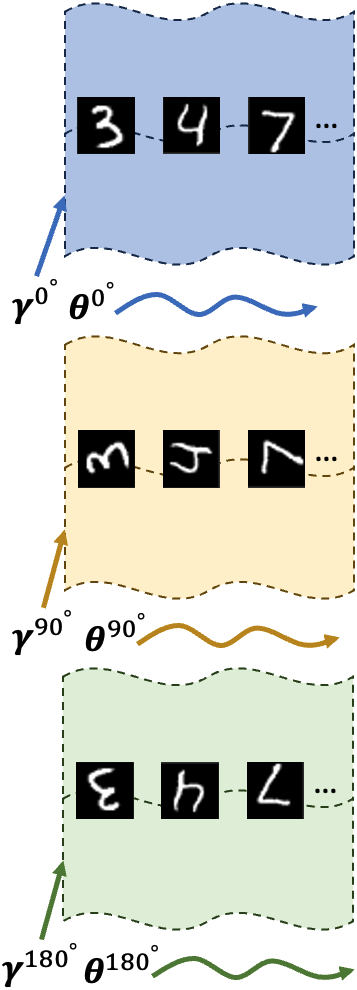}
\caption{Data generating process.}
\label{fig:generating_process}
\end{subfigure}
\hfill
\begin{subfigure}[t]{0.84\textwidth}\centering
\includegraphics[width=\textwidth]{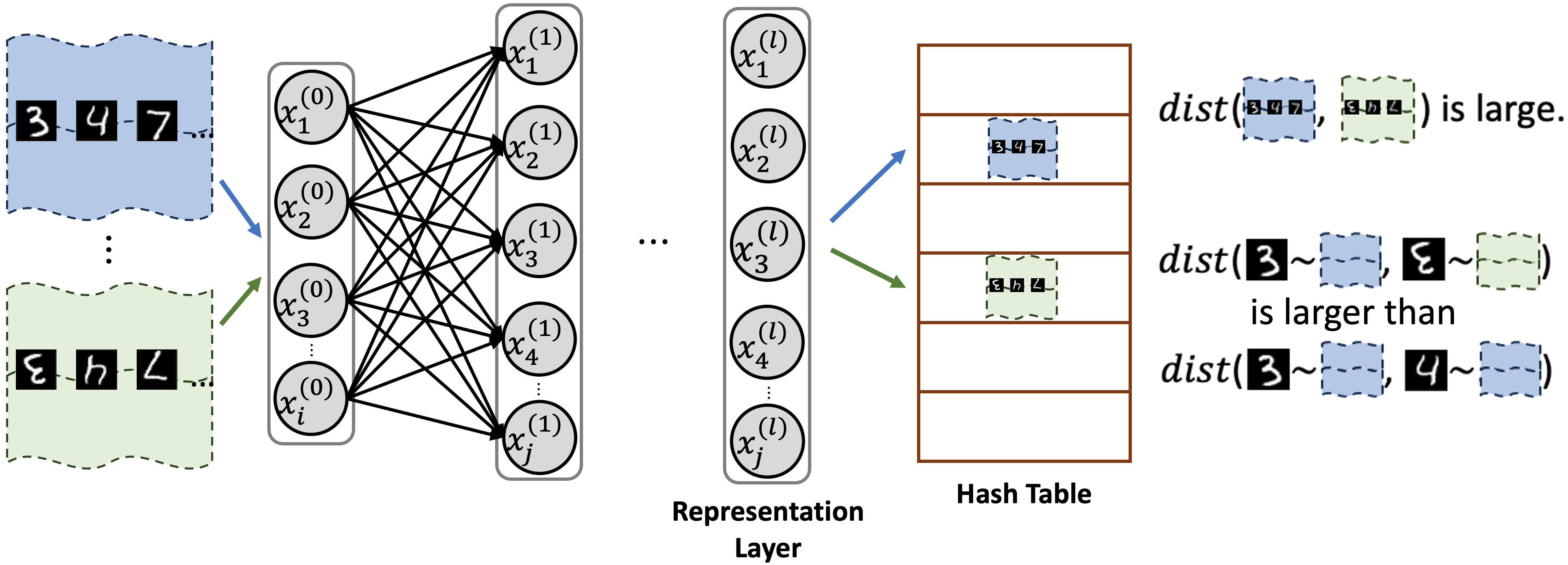}
\caption{A DNN as a Task-specific Geometry Sensitive Hash (T-GSH) function.}
\label{fig:task_specific_geometric_hash}
\end{subfigure}
\hfill
\label{fig:overall_ideas}
\caption{(a)~Each task $t$ consists of a set of points of classes on a simple manifold. Furthermore, each point in the task can be characterized by three latent parameters: $\boldsymbol{\gamma}^{t}, \boldsymbol{\delta}, \boldsymbol{\theta}^{t}$. Notably, we consider a task-specific manifold broader than the class-level one in~\cite{dikkala2021manifold}. (b)~Any set of points on the same task manifold map to (approximately) the same representation (i.e., the penultimate layer feature map), while a set of points from different task manifolds go to far away representations.}
\end{figure*}

\section{Task-Specific Geometric Sensitive Hashing}
\label{sec:t_gsh}

\subsection{Problem Definition and Proposed Randomly Weighted Neuromodulation Method}

Following~\cite{hong2022learning}, we consider a supervised continual-learning setup where $\mathcal{T}$ tasks arrive to a learner in sequential order. Let $\mathcal{D}_{t} = \{ \mathbf{x}_{i, t}, y_{i,t} \}_{t}^{n_{t}}$ be the dataset of task $t$, composed of $n_{t}$ pairs of raw instances and corresponding labels. For simplicity, we adopt a simple task-incremental continual learning setup where $\mathcal{C}$ is the number of classes for every task. Next, we extend the manifolds from~\citet{dikkala2021manifold} to represent \emph{task-specific} manifolds. We assume that each input $\mathbf{x}_{i, t} \in \mathbb{R}^{d}$ is drawn from a mixture distribution over $\mathcal{T}$ manifolds $M_{1},\dots, M_{\mathcal{T}}$ sharing similar topologies~(Fig.~\ref{fig:generating_process}), and each label $y_{i, t} \in [\mathcal{C}]$ corresponds to the manifold of $\mathbf{x}_{i, t}$. Moreover, each point $\mathbf{x}_{t}$ on manifold $M_{t}$ is determined by the latent vectors, $\boldsymbol{\gamma}^{t} \in \mathbb{R}^{s}$, $\boldsymbol{\delta} \in \mathbb{R}^{p}$, and  $\boldsymbol{\theta}^{t} \in \mathbb{R}^{k}$, where:
\begin{itemize}
    \item $\boldsymbol{\gamma}^{t}$ is the manifold identifier for the task $t$, and there is a one-to-one correspondence between $\boldsymbol{\gamma}^{t}$ and $M_{t}$.
    \item $\boldsymbol{\delta}$ is a set of manifold identifiers for classes. This is an explicitly shared manifold, representing canonical representation objects and is shared across all task manifolds $\boldsymbol{\gamma}$.
    \item $\boldsymbol{\theta}^{t}$ is the transformation in the task $t$. So, if we fix $\boldsymbol{\gamma}^{t}$, the manifold $M_t$ can be generated by sampling different values of  $\boldsymbol{\theta}^{t}$.
\end{itemize}

Intuitively, $\boldsymbol{\gamma}^{t}$ represents the centroid of a cluster for the task $t$, $\boldsymbol{\delta}$ includes explicitly shared clusters for labels that every $\boldsymbol{\gamma}$ contains, and $\boldsymbol{\theta}^{t}$ represents a small perturbation around the centroid. So, the manifold $M_{\boldsymbol{\gamma}^{t}}$ is comprised of all inputs $\mathbf{x}_{t}$ of the form $\{ \boldsymbol{\gamma}^{t} + \boldsymbol{\delta} + \boldsymbol{\theta}^{t} |\; \lVert \boldsymbol{\theta}^{t} \rVert \leq \epsilon \}$. In this setting, we employ a Locality Sensitive Hashing~(LSH) to map each input to a hash bucket. With appropriate configuration, we empirically demonstrate that DNNs exhibit LSH-like behavior on the family of manifolds with a shared geometry defined by a set of analytic functions. In particular, extending the work of \citet{dikkala2021manifold}, we show that the penultimate ``representation layer''~$r$~(Fig.~\ref{fig:task_specific_geometric_hash}) of an appropriately trained network will satisfy the following property:
\begin{define}[Task-specific Geometric Sensitive Hashing~(T-GSH)]
We say $r$ is a T-GSH function with respect to a set of task manifolds if:
    \begin{enumerate}[a)]
        \item For any two points on the same task manifold $x_{1, t}, x_{2, t} \in M_{t}$, the distance $\lVert r(x_{1, t}) - r(x_{2, t}) \rVert$ is small regardless of the associated classes.
        \item For any two points on two well-separated task manifolds $x_{1, t} \in M_{t} \;\text{and}\; x_{2, t'} \in M_{t'}, \lVert r(x_{1, t}) - r(x_{2, t'}) \rVert$ is large regardless of the associated classes.
    \end{enumerate}
\label{def:tgsh}
\end{define}
%
That is, DNNs whose representations satisfy the T-GSH properly capture the shared task manifold geometry similar to how LSH functions capture spatial locality. Finally, by employing the recoverability of the manifold~\cite{dikkala2021manifold}, we give empirical evidence to support the recoverability of $\boldsymbol{\gamma}^{t}$ via simple transformations on top of representations learned by DNNs which behave as T-GSH functions.

We consider task manifolds as subsets of points in $\mathbb{R}^{d}$. Every manifold $M_{\boldsymbol{\gamma}^{t}}$ has an associated latent vector $\boldsymbol{\gamma}^{t} \in \mathbb{R}^{s}$ (where $s \leq d$) which acts as an identifier of $M_{\boldsymbol{\gamma}^{t}}$. The task manifold is then defined to be the set of points $\mathbf{x}_{t} = \mathbf{f}(\boldsymbol{\gamma}^{t}, \boldsymbol{\delta}, \boldsymbol{\theta}^{t}) = (f_{1}(\boldsymbol{\gamma}^{t}, \boldsymbol{\delta}, \boldsymbol{\theta}^{t}), \dots, f_{d}(\boldsymbol{\gamma}^{t}, \boldsymbol{\delta}, \boldsymbol{\theta}^{t})) \;\text{for}\; \boldsymbol{\delta} \in \Delta \subseteq \mathbb{R}^{p}, p < d, \; \text{and for} \; \boldsymbol{\theta}^{t} \in \Theta \subseteq \mathbb{R}^{k}, k < d$. Here, the manifold generating function $\mathbf{f} = \{ f_{i}(\cdot, \cdot) \}_{i=1}^{d}$ where the $f_{i}$ are all analytic functions. Without significant loss of generality, we assume our inputs $\mathbf{x}$ and $\boldsymbol{\gamma}$ are normalized and lie on $S^{d-1}, $ and $S^{s-1}$, the $d$ and $s$-dimensional unit spheres, respectively. Given the above generative process, we leverage the Assumption~1, which allows us to use a special form of the DNN configuration used by \citet{dikkala2021manifold}.
\begin{assum}[Modified Invertibility~\cite{dikkala2021manifold}]
    There is an analytic function $g(\cdot): \mathbb{R}^{d} \rightarrow \mathbb{R}^{s}$ with bounded norm Taylor expansion such that for every point $\mathbf{x}_{t} = \mathbf{f}(\boldsymbol{\gamma}^{t}, \boldsymbol{\delta}, \boldsymbol{\theta}^{t}) \;\text{on}\; M_{\boldsymbol{\gamma}^{t}}, g(\mathbf{x}_{t}) = \boldsymbol{\gamma}^{t}$.
\label{assum:invert}
\end{assum}
%

The conventional GSH neural network~\cite{dikkala2021manifold} is defined as a single-hidden-layer network, $\mathbf{y} = A \cdot B \cdot \sigma(C \mathbf{x})$, where the input $\mathbf{x} \in \mathbb{R}^{d}$ passes through a wide, randomly initialized, fully connected, non-trainable layer $C \in \mathbb{R}^{D \times d}$ followed by a ReLU activation $\sigma(\cdot)$. Then, there are two trainable, fully connected layers $A \in \mathbb{R}^{\mathcal{C} \times T}$, $B \in \mathbb{R}^{T \times D}$ with no non-linearity between them. Following the above configuration, we define a special-case T-GSH neural network as follows:
\begin{equation}
    \hat{\mathbf{y}}_{t} = \mathbf{R} \cdot B^{t} \cdot \sigma(C \mathbf{x})
    \label{eq:nn_tgsh}
\end{equation}
for each task $t$. Thus, not only does a T-GSH differ from a GSH by the introduction of multiple task labels, but the learnable matrix $A$ from a standard GSH is replaced with another randomly weighted matrix $\mathbf{R}$ that acts as the explicitly shared manifold $\boldsymbol{\delta}$ in our definition (i.e., it is shared across all tasks). Intuitively, a randomly weighted matrix is nearly orthogonal, and so learning task-specific $B^{t}$ must be enforced to learn the commonality in the task $t$. In the upcoming section, we connect a T-GSH neural network with one of the examples of neuromodulation-inspired DNNs for continual learning.



\subsection{Revising Configurable Random Weight Networks}
\label{sec:crwn}
Configurable Random Weight Networks~(CRWNs)~\cite{hong2022learning} were proposed as a simple yet effective form of artificial neuromodulatory systems for continual learning. The proposed approach consists of two types of artificial neuromodulation, \emph{Global} and \emph{Local} modulation, and shows that their neuromodulatory signals learn how to adjust long-lasting/unchanging random synaptic weights for specific tasks, enabling task-specific learning. Furthermore, the proposed neuromodulation can be applied in a layer-wise fashion. To improve the efficiency of training and the applicability to any network architecture, \citet{hong2022learning} proposed two kinds of model architectures, which we summarize in the single expression in Eq.~\eqref{eq:crwn},
\begin{equation}
\begin{split}
    \hat{\mathbf{y}}_{t} &= \alpha_{t} \cdot \mathbf{R} \cdot (v^{t} \odot \sigma(C \mathbf{x})) \\
    &= \mathbf{R} \cdot ((\alpha_{t} \cdot v^{t}) \odot \sigma(C \mathbf{x}))
\end{split}
    \label{eq:crwn}
\end{equation}
where:
\begin{itemize}
    \item $\mathbf{R} \in \mathbb{R}^{\mathcal{C} \times D}$ and $C \in \mathbb{R}^{D \times d}$ are non-trainable, randomly weighted matrices.
    \item $\alpha_{t} \in \mathbb{R}$ is a learnable constant acting as \emph{global} neuromodulation.
    \item $v^{t} \in \mathbb{R}^{D}$ is a learnable vector mimicking \emph{local} neuromodulation.
\end{itemize}
In the following, we view Eq.~\eqref{eq:crwn} through the lens of T-GSH in Eq.~\eqref{eq:nn_tgsh}.

\paragraph{CRWN as a T-GSH function.}
CRWNs were first proposed as an efficient CL method by highlighting the approach's benefits concerning computational efficiency. However, we highlight here that CRWNs are a T-GSH function achieved by the proposed artificial neuromodulatory systems and thus can be used to shed light on the manifold learning facilitated by neuromodulatory processes. Below, we compare Eq.~\eqref{eq:nn_tgsh} and \eqref{eq:crwn}.
\begin{equation*}
    \hat{\mathbf{y}}_{t} =
    \mathord{\overbrace{\mathbf{R} \cdot B^{t} \cdot \sigma(C \mathbf{x})}^{\text{\protect{Eq.~\eqref{eq:nn_tgsh}}}}}
    =
    \mathord{\overbrace{\mathbf{R} \cdot \mathord{(\underbrace{(\alpha_{t} \cdot v^{t})}_{B^{t}}} \odot \sigma(C \mathbf{x}))}^{\text{\protect{Eq.~\eqref{eq:crwn}}}}}
    \label{eq:comparision_two_eqs}
\end{equation*}
Thus, $\alpha_{t} \cdot v^{t}$ from the CRWN in Eq.~\eqref{eq:crwn} plays the role of $B^{t}$ in the T-GSH in Eq.~\eqref{eq:nn_tgsh}, and so a CRWN is a simple realization of a T-GSH. Following \citet{dikkala2021manifold}, it is important to use a proper regularization term on the weights $B^{t}$ in the classification loss for training a T-GSH function, and the learnable constant $\alpha_{t}$ in CRWNs acts as a similar regularizer for $v^{t}$. 

\section{Experiments}
\label{sec:exp}
\begin{figure*}\centering%
    \begin{subfigure}[t]{0.38\textwidth}\centering
        \includegraphics[width=\textwidth]{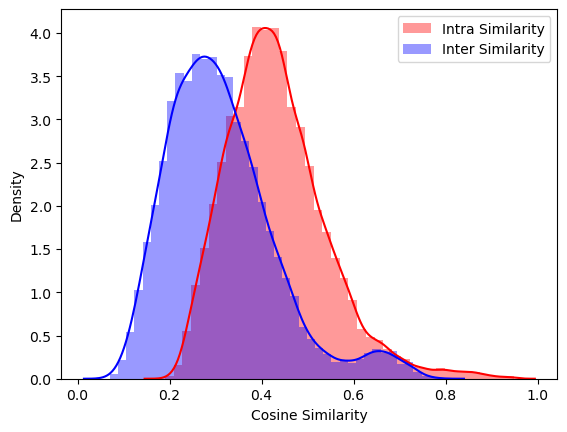}
        \caption{A comparison of cosine similarities of the points. (\textbf{Intra similarity}): the cosine similarities of the points with the \emph{different labels} on the \emph{same task manifold}. (\textbf{Inter similarity}): the cosine similarities of the points with the \emph{same label} on the \emph{different task manifolds}.}
        \label{fig:cosine_distance}
    \end{subfigure}
    \hfill
    \begin{subfigure}[t]{0.6\textwidth}\centering
        \includegraphics[width=\textwidth]{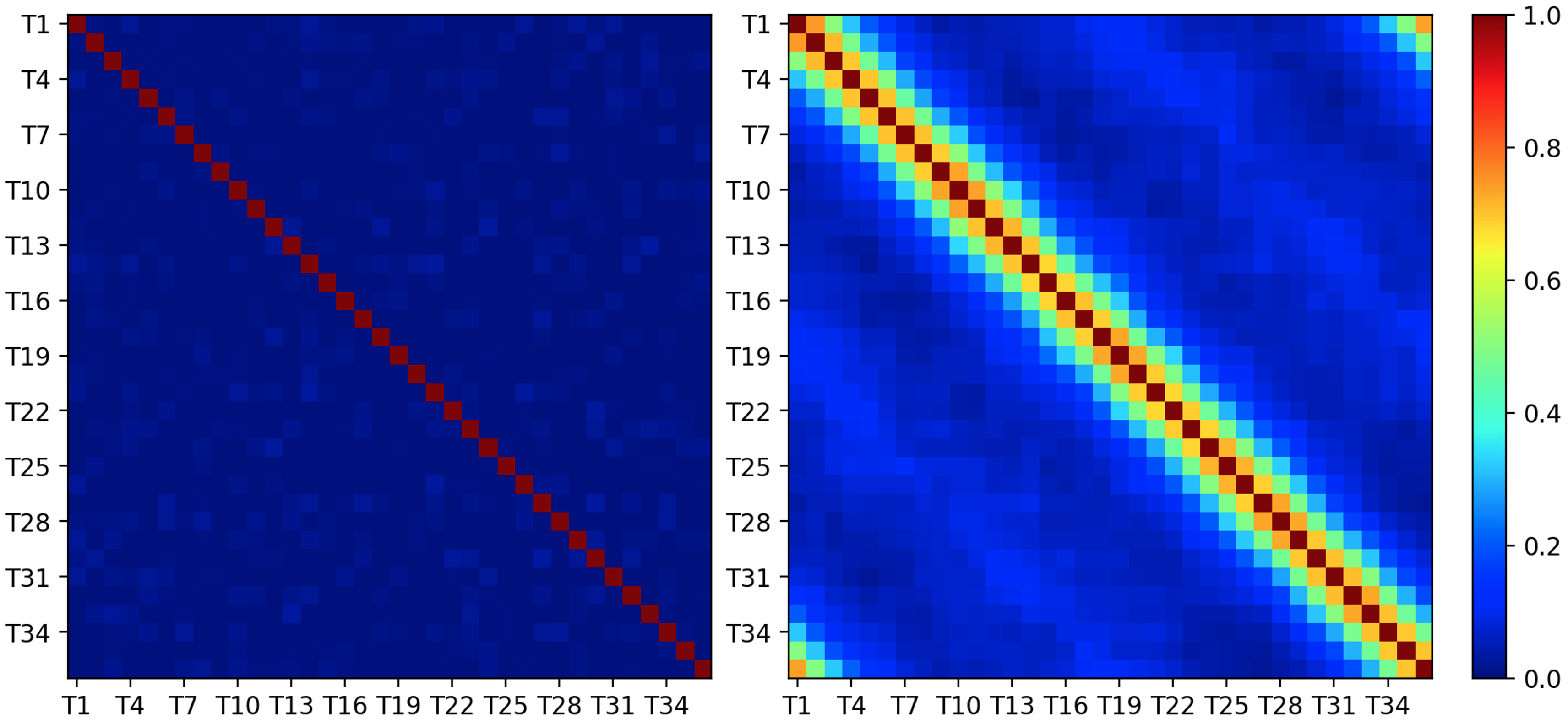}
        \caption{A confusion matrix of intra~(same task manifold) VS inter~(different task manifolds) cosine similarity of task representations trained on \emph{RotationMNIST}. Cosine similarity between task context vectors before~(\textbf{left}) and after training~(\textbf{right}). Notably, adjacent tasks are expected to be the most similar to each other as they represent the smallest angular difference in images.}
        \label{fig:rmnist_cosine_sim}
    \end{subfigure}
    \caption{Experiment results on RotationMNIST.}
    \label{fig:exp_rmnist}
\end{figure*}
In this section, we demonstrate the CRWN is a T-GSH function using variants of the MNIST~\cite{lecun1998mnist} dataset, including \emph{RotationMNIST}, and our newly proposed experimental setup, \emph{ShiftMNIST} and \emph{AugmentMNIST}. 

\subsection{Experiments for showing that CRWN is a T-GSH function.}

We first validate the CRWN on RotationMNIST to show it is a T-GSH function. RotationMNIST is one of the popular CL benchmark datasets. Following \citet{hong2022learning}, we define the dataset so that each of the 36 tasks corresponds to images counterclockwise rotated by a multiple of 10 degrees.

\paragraph{CRWNs as a GSH and T-GSH function.}
Using experimental results of CRWNs for continual learning, we can empirically demonstrate that CRWNs are T-GSH functions. In particular, \citet[Table 1]{hong2022learning} showed that CRWNs achieved about $95\%$ test accuracy average over all 36 RotationMNIST tasks~(\texttt{FlyNet}: $94.9\%$ and \texttt{NeuroModNet}: $95.5\%$, respectively). This indicates that CRWNs perform well as a GSH function that successfully classifies all labels for every task.

Figure~\ref{fig:cosine_distance} depicts a comparison of cosine similarities of latent CRWN task representations. \emph{Intra similarity} indicates the cosine similarities of the points with different labels on the same task manifold (i.e., same RotationMNIST rotation angle). In contrast, \emph{Inter similarity} means the cosine similarity of the points with the same label, but on the different task manifolds (i.e., different RotationMNIST rotational angles). This shows CRWN is a T-GSH function in Fig.~\ref{fig:task_specific_geometric_hash}, showing that the similarity of the points having different labels, but on the same task manifold must be larger than one of the points with the same label, but on the different task manifolds.

\paragraph{Satisfaction of Assumption~\ref{assum:invert}.}
\begin{figure}[t!]
    \centering
    \includegraphics[width=0.7\textwidth]{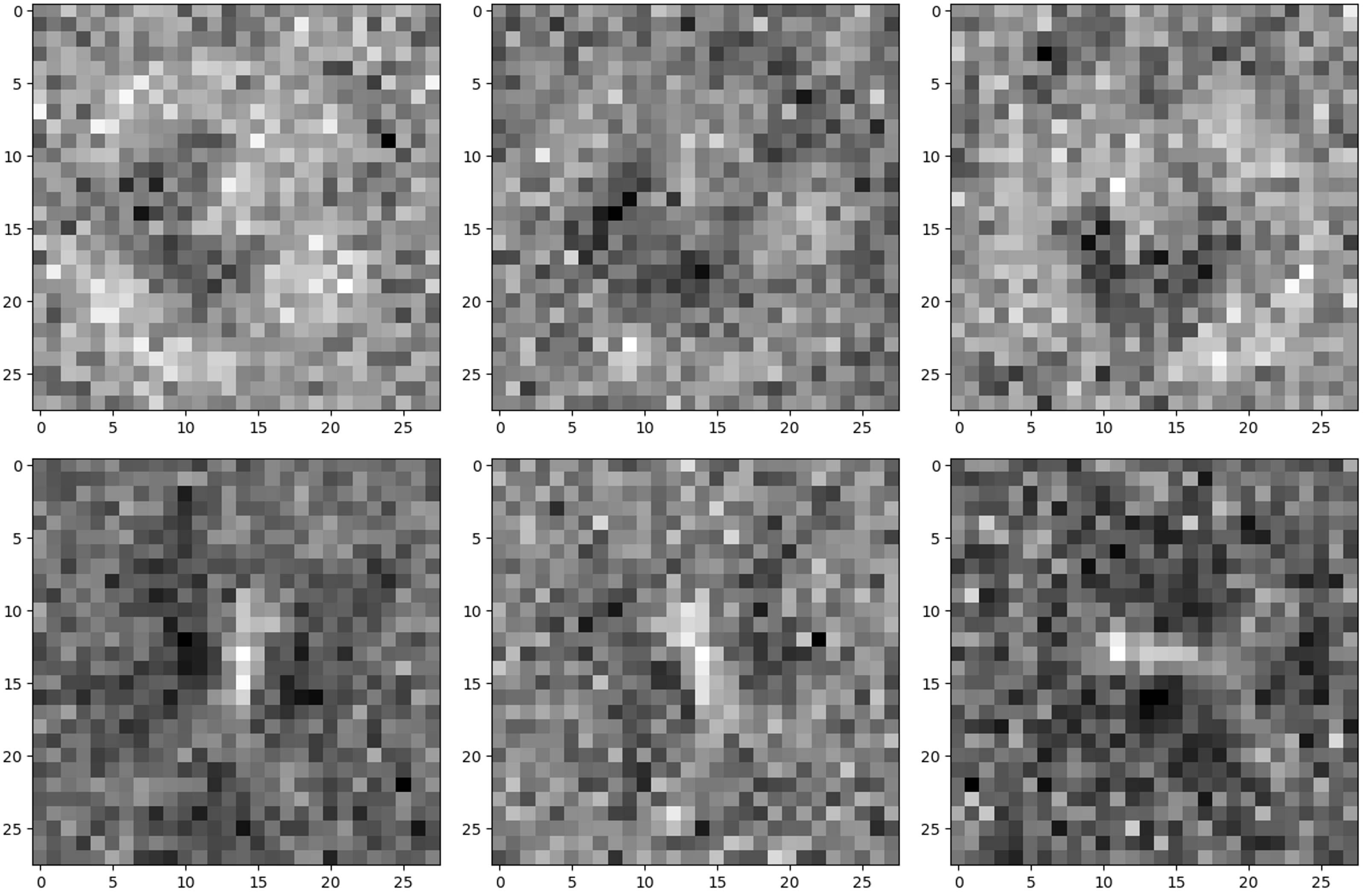}
    \caption{Empirical evidence of approximate data generating process with our defined invertibility assumption on RotationMNIST. \textbf{(Top Row)} Reconstructed samples of digit ``3''. \textbf{(Bottom Row)} Reconstructed samples of digit ``1''. \textbf{(Left Column)} Reconstructed samples of digits from the task manifold $M^{T1}$, which is $0$\degree rotation. \textbf{(Middle Column)} the samples of digits from the task manifold $M^{T5}$, which is $40$ \degree counterclockwise rotation. \textbf{(Right Column)} the samples of digits from the task manifold $M^{T10}$, which is $90$ \degree counterclockwise rotation.}
    \label{fig:recon_digits_rotation}
\end{figure}
We can also show that the inverse function of CRWNs can mimic the data-generating process, satisfying our Assumption~\ref{assum:invert}. Figure~\ref{fig:recon_digits_rotation} depicts the reconstructed image samples from CRWNs from the learned task manifolds.

Because of the ReLU function, we cannot obtain the complete inverse function of CRWNs. Thus, we omitted ReLU and computed the pseudo-inverse of the weights of the learned CRWNs. Because we fixed the manifolds for classes $\boldsymbol{\delta}$ using the non-trainable matrix $\mathbf{R}$, each row in $\mathbf{R}$ can be seen as the canonical representations of objects. As shown in Fig.~\ref{fig:recon_digits_rotation}, the canonical representation for the digit ``3'' is the 4th row, and for the digit ``1'' is the 2nd row. Thus, let $\mathbf{f}$ be the inverse CRWNs omitting the ReLU function. We can reconstruct a digit sample with class $i$ as follows:
\begin{equation}
    \Bar{\mathbf{x}}_{i, t} = \mathbf{f}(\textbf{s}) \;\text{where}\; \mathbf{s} \sim \mathcal{N}(\mu_{i,t}, \sigma_{i, t}), \mu_{i, t} = R_{i}^{\top} \odot (\alpha_{t} \cdot v_{t}), \sigma_{i, t} = \mathbf{1}_{D} \cdot 1/D 
\end{equation}
$R_{i}$ is $i$-th row of the matrix $\mathbf{R}$ and $\mathbf{1}_{D}$ is the vector containing ones with the size $D$. In Fig.~\ref{fig:recon_digits_rotation}, despite approximate reconstruction, the shape of digits and the rotation can be shown, indicating the CRWN satisfies Assumption~\ref{assum:invert}.

\subsection{Experiments For Task Similarities}
In this section, due to the architectural characteristic of a T-GSH function, we demonstrate unique features of the CRWN that allow for computing the relationship among tasks. In particular, we introduce the scalar--vector product $c_{t} \triangleq \alpha_{t} \cdot v^{t}$ so that the CRWN expression becomes:
\begin{equation*}
    \hat{\textbf{y}}_{t} = \mathbf{R} \cdot ((\alpha_{t} \cdot v^{t}) \odot \sigma(C \mathbf{x})) = \mathbf{R}(c_{t} \odot \sigma(C \mathbf{x})).
\end{equation*}
We refer to $c_{t}$ as a \emph{context vector.} The following experiments on RotationMNIST, ShiftMNIST, and AugmentMNIST demonstrate the semantic interpretation of the context vector $c_{t}$ among tasks.

\paragraph{RotationMNIST.}
Figure~\ref{fig:rmnist_cosine_sim} shows the pairwise cosine similarity\footnote{We compute $1-\texttt{scipy.spatial.distance.cosine}$.} between context-vector pairs $(c_{j},c_{k})$ for every $(j,k)$ pair of RotationMNIST tasks, shown both before training and after training. In the task ordering, adjacent tasks are expected to be the most similar to each other as they represent the smallest angular difference in images. In other words, networks trained for task~$i$ should have minimal degradation in performance for test data from tasks $i-1$ and $i+1$. Consistent with this functional expectation, the mechanistic pattern Fig.~\ref{fig:rmnist_cosine_sim} shows greatest similarity between adjacent context vectors after training, and so trained context vectors exist in a representational space that captures fundamental similarities among tasks. 

\paragraph{ShiftMNIST.}
\begin{figure}\centering
    \begin{subfigure}[t]{0.39\textwidth}\centering
        \includegraphics[width=\textwidth]{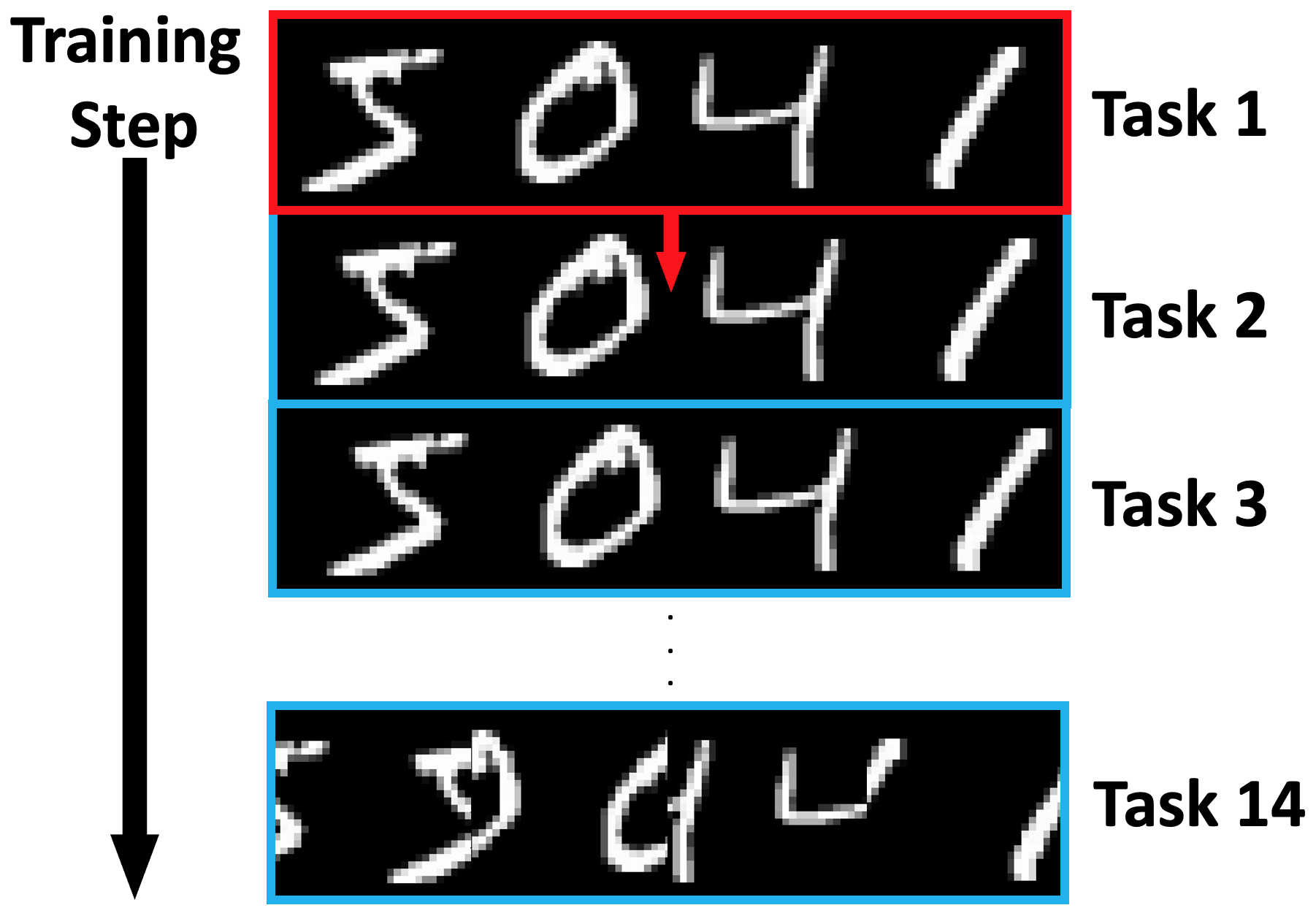}
        \caption{ShiftMNIST-1px tasks}
        \label{fig:shiftmnist_task_01}
    \end{subfigure}
    \begin{subfigure}[t]{0.59\textwidth}\centering
        \includegraphics[width=\textwidth]{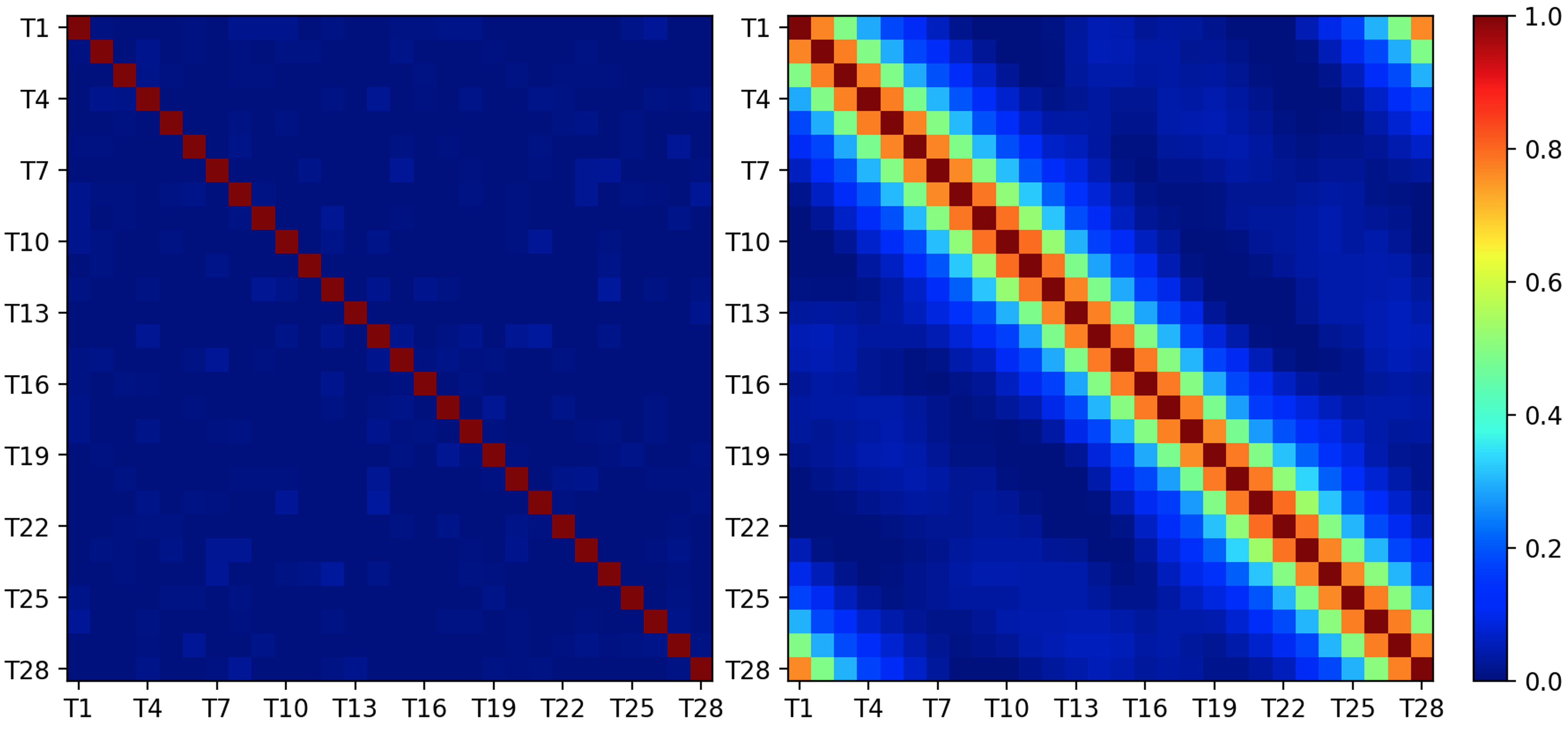}
        \caption{ShiftMNIST-1px results}
        \label{fig:shiftmnist_cosinesim_01}
    \end{subfigure}
    \caption{ShiftMNIST-1px. (\subref{fig:shiftmnist_task_01}) The difference between the datasets of the two adjacent tasks, $i$ and $i+1$, is that the images of task $i+1$ are the images of task $i$ shifted a single pixel to the right. Thus, the total number of tasks to learn is 28. (\subref{fig:shiftmnist_cosinesim_01}) Cosine similarity between the context vectors for each task before (\textbf{left}) and after training (\textbf{right}).}
\end{figure}
\begin{figure}\centering
    \begin{subfigure}[t]{0.39\textwidth}\centering
        \includegraphics[width=\textwidth]{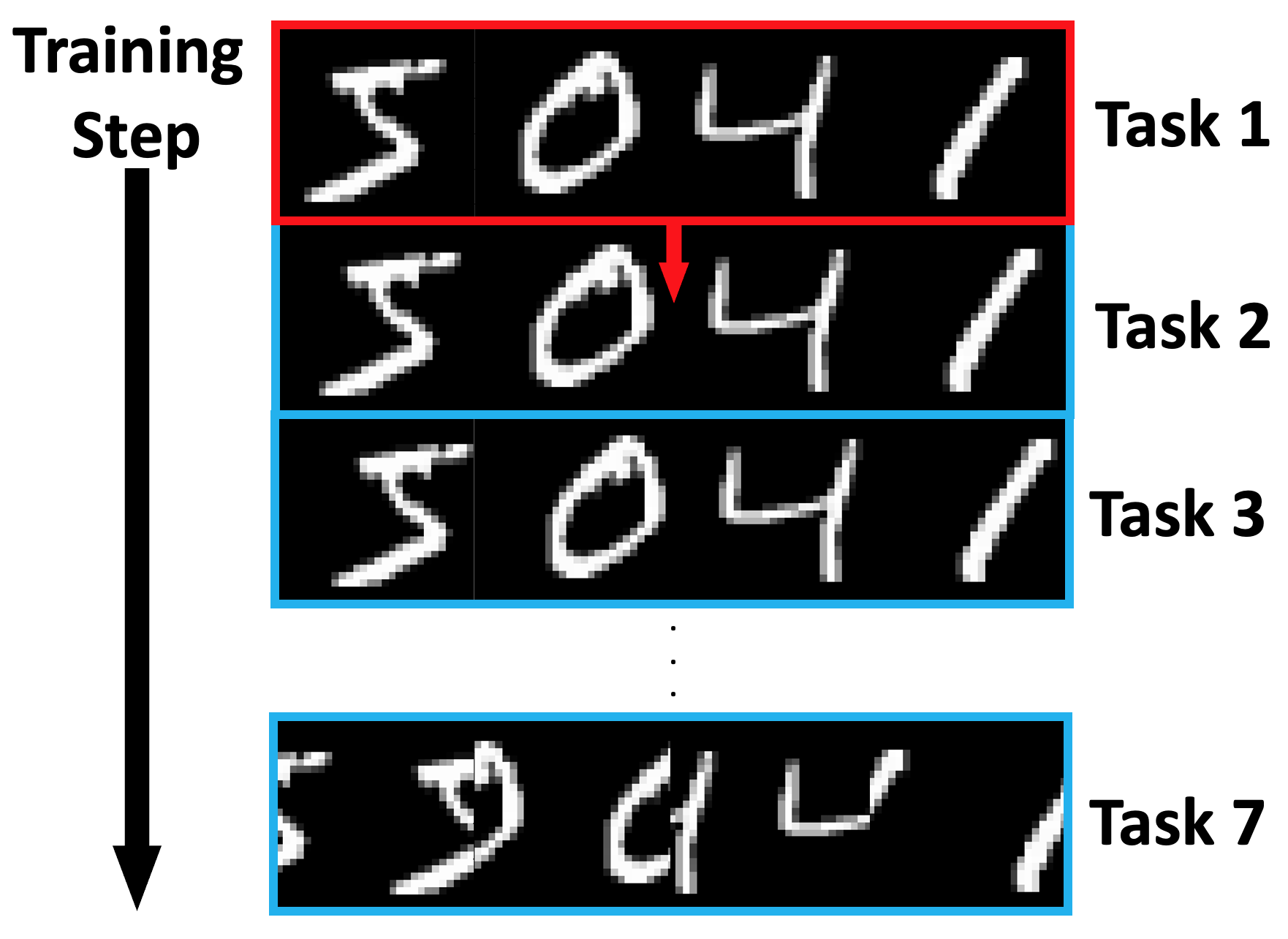}
        \caption{ShiftMNIST-2px tasks}
        \label{fig:shiftmnist_task_02}
    \end{subfigure}
    \begin{subfigure}[t]{0.59\textwidth}\centering
        \includegraphics[width=\textwidth]{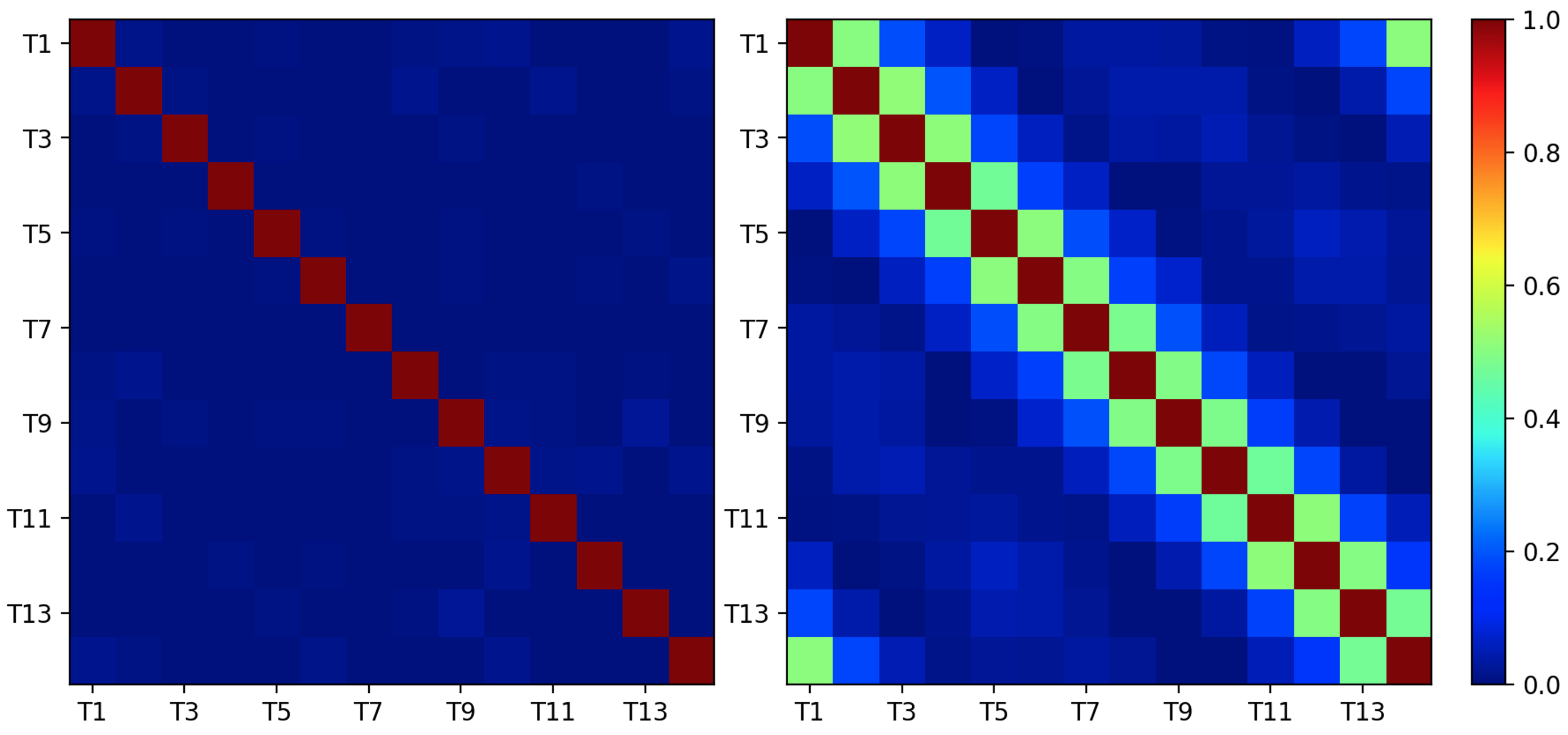}
        \caption{ShiftMNIST-2px results}
        \label{fig:shiftmnist_cosinesim_02}
    \end{subfigure}
\caption{ShiftMNIST-2px. (\subref{fig:shiftmnist_task_02}) The difference between the datasets of the two adjacent tasks, $i$ and $i+1$, is that the images of task $i+1$ are the images of task $i$ shifted two pixels to the right. Thus, the total number of tasks to learn is 14. (\subref{fig:shiftmnist_cosinesim_02}) Cosine similarity between the context vectors for each task before (\textbf{left}) and after training (\textbf{right}).}
\end{figure}
To test whether geometric adjacency generally implies similarity of trained vectors, we introduce a new sequence of tasks, \emph{ShiftMNIST}, as shown in Figs~\ref{fig:shiftmnist_task_01} and~\ref{fig:shiftmnist_task_02}. Whereas RotationMNIST rotates each of the MNIST characters, ShiftMNIST translates them laterally by a number of pixels (with periodic boundaries). Adjacent ShiftMNIST-1px tasks in Fig.~\ref{fig:shiftmnist_task_01} are separated by 1~pixel, and so an image in Task~3 is the same as an image in Task~2 shifted to the right by 1~pixel. Similarly, adjacent ShiftMNIST-2px tasks in Fig.~\ref{fig:shiftmnist_task_02} are separated by 2 pixels. Because MNIST images are each 28 pixels wide, there are 28 distinct ShiftMNIST-1px tasks and 14 distinct ShiftMNIST-2px tasks. We conduct two experimental setups separately using different random seeds and check whether meaningful relationships with each other can be derived.
The cosine similarities of context vectors before and after training for ShiftMNIST-1px and ShiftMNIST-2px are shown in Figs~\ref{fig:shiftmnist_cosinesim_01} and~\ref{fig:shiftmnist_cosinesim_02}, respectively.
The diagonal pattern in both cases matches that of Fig.~\ref{fig:rmnist_cosine_sim}, which confirms that the similarity vectors for ShiftMNIST have encoded geometric relationships among tasks in a similar way as with RotationMNIST. Furthermore, Fig.~\ref{fig:shiftmnist_cosinesim_02} appears as a more coarsely subsampled version of Fig.~\ref{fig:shiftmnist_cosinesim_01}, which demonstrates that cosine similarity differences are repeatable and a function of the tasks themselves and not an artifact of the architecture of the network. Thus, training these compact context vectors extracts meaningful semantic information about individual tasks.

\begin{figure}[t!]
\centering
\includegraphics[width=0.65\textwidth]{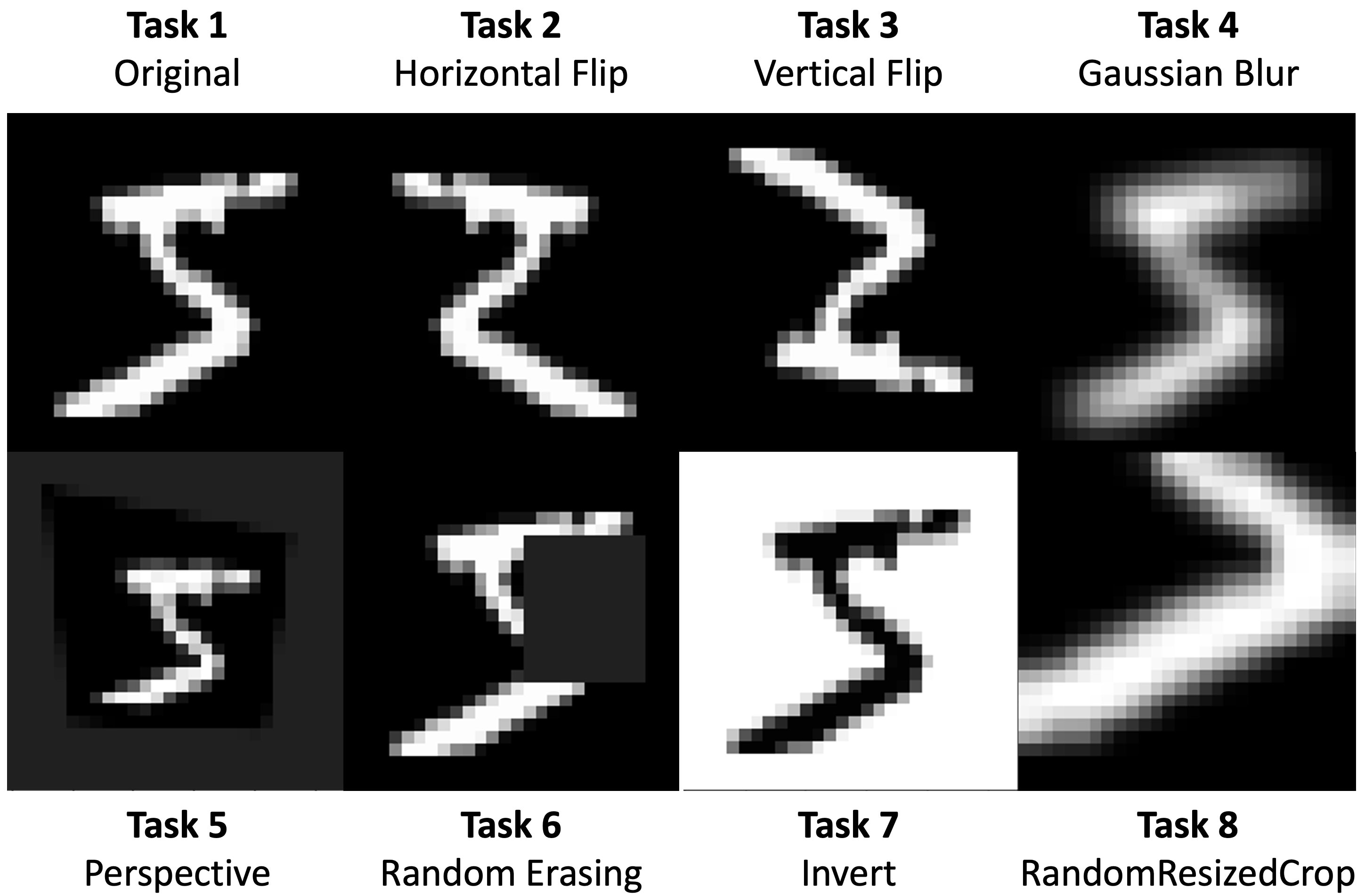}
\caption{AugmentMNIST task. The list of variants of augmentation methods applied to produce the task. See the implementation details in Table~\ref{tab:pytorch_augmentmnist}.}
\label{fig:augmentmnist_task}
\end{figure}

\begin{figure}[t!]
\centering
\includegraphics[width=0.9\textwidth]{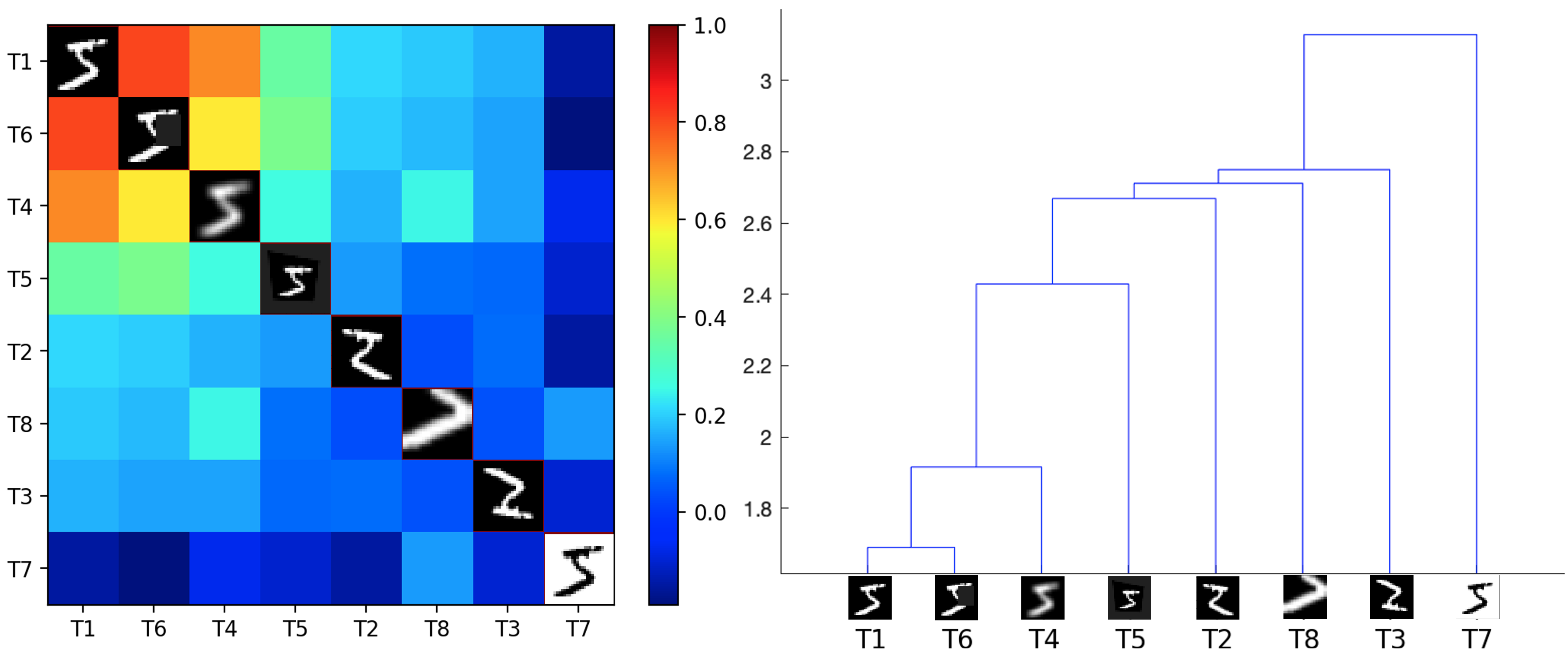}
\caption{AugmentMNIST results. Sorted cosine similarity between the context vectors for each task after training (\textbf{left}) and task-similarity clustering~(\textbf{right}).}
\label{fig:augmentmnist_consinesim_2}
\end{figure}

\paragraph{AugmentMNIST.}
Here, we propose AugmentMNIST, which employs a sequence of 8 off-the-shelf, commonly used data-augmentation tasks as shown in Fig.~\ref{fig:augmentmnist_task}~(e.g., Random Erasing~\cite{zhong2020random} that is provided by PyTorch). We list the details of the setting in Table~\ref{tab:pytorch_augmentmnist}. After training on each of the 8 tasks, we use hierarchical agglomerative clustering to sort the tasks so that adjacent tasks tend to have highest similarity.
\begin{table*}\centering
\caption{PyTorch routines used for AugmentMNIST. Here, \texttt{TT} is shorthand for \texttt{torchvision.transforms}.}
\label{tab:pytorch_augmentmnist}
\begin{tabular}{r@{:\quad}l}
    \toprule
    \textbf{AugmentMNIST Task} & \textbf{PyTorch method and configuration}\\
    \midrule
    \midrule
    Horizontal Flip~(T2) & \texttt{TT.RandomHorizontalFlip(p=1.0)} \\
    Vertical Flip~(T3) & \texttt{TT.RandomVerticalFlip(p=1.0)} \\
    Gaussian Blur~(T4) & \texttt{TT.GaussianBlur(kernel\_size=5, sigma=2)} \\
    Perspective~(T5) & \texttt{TT.RandomPerspective(distortion\_scale=0.5, p=1.0)} \\
    Random Erasing~(T6) & \texttt{TT.RandomErasing(p=1.0)} \\
    Invert~(T7) & \texttt{TT.RandomInvert(p=1.0)}\\
    RandomResizedCrop~(T8) & \texttt{TT.RandomResizedCrop(size=28)}\\
    \bottomrule
\end{tabular}
\end{table*}
Fig.~\ref{fig:augmentmnist_consinesim_2} shows the resulting cosine similarities as well as a dendrogram of the hierarchical clustering by cosine similarity.
%
The results show that context-vector representations of Random Erasing~(T6), Gaussian Blur~(T4), and Perspective~(T5) are very close to that of the original image~(T1); in other words, the four tasks that maintain the shape, color, and orientation of the image are grouped together. Furthermore, the inverted-image task~(T7) is an outgroup as it has negative cosine similarity with all other tasks, which is consistent with it being the only task operating on a white background. Furthermore, the cosine similarities suggest a closer relationship between T1 and T7 than between T2 and T7, indicating that the horizontal flip combined with the color inversion has an additive effect in terms of context vector dissimilarity. Thus, the geometry of the trained context-vector space encodes semantic information about comparisons between tasks themselves.

%

%

\section{Related Work}
\label{sec:related_work}
There have been various works on studying classification problems as manifold learning~\cite{tenenbaum2000global, belkin2006manifold}. Furthermore, manifold learning has been adapted in various literature, such as feature learning with different modalities~\cite{nguyen2022deep}, explainability~\cite{han2022enhance}, speech recognition~\cite{tomar2013efficient} and neuroscience~\cite{busch2023multi}. We leverage the concept of Locality Sensitive Hashing, and this property has been extensively studied in insect-brain-like architecture~\cite{dasgupta2017neural}, and several works have been proposed in novelty detection~\cite{dasgupta2018neural} and relational learning~\cite{hong2021insect, hong2022representing}. The connection between DNNs and Hash Functions was explored before~\cite{wang2017survey}. Another related work is the benefits of wide non-linear layers~\cite{daniely2016toward} and over-parameterized networks~\cite{arora2019fine, allen2019learning}.  We empirically demonstrate that T-GSH holds for certain architectures under the manifold data assumption given a series of classification tasks.

\section{Discussion}
\label{sec:discussion}
We studied the problem of a sequence of multiple supervised classification tasks as a task manifold-learning problem under a specific generative process wherein the manifolds share geometry. We demonstrated empirically  that properly trained DNNs with a neuromodulation-inspired architecture satisfy the T-GSH property by recovering each task's semantically meaningful latent representation $\gamma$. Our work provides new geometric intuitions about the functioning of randomly weighted neuromodulation systems found throughout nature and allows us to better understand how randomness helps exploit finite representation spaces. We plan to extend our work to provide theoretical evidence in various continual-learning setups, such as class-incremental and domain-incremental learning.



\section*{Acknowledgements}

This work was supported in part by NSF award 2223839 and USACE ERDC award W912HZ-21-2-0040.

\bibliography{main}
\bibliographystyle{abbrvnat}

\newpage

\end{document}